\documentclass[11pt,letterpaper]{article}
\usepackage{emnlp2017-titleboxchange}
\usepackage{times}
\usepackage{latexsym}
\usepackage[fleqn]{amsmath}
\usepackage{url}
\usepackage{booktabs}
\usepackage{multirow}
\usepackage{graphicx}
\usepackage{caption}
\usepackage{subcaption}
\usepackage{mathabx}
\usepackage{xparse}

\usepackage{xspace}

\usepackage{mathpartir}
\usepackage{tikz}
\usepackage{tikz-dependency}

\def\namecite{\newcite}

\usepackage{etoolbox}

\usepackage{array,booktabs,ragged2e} %
\usepackage{amsthm}
\usepackage{setspace}
\newcolumntype{R}[1]{>{\RaggedLeft\arraybackslash}p{#1}}

\emnlpfinalcopy

\def\emnlppaperid{1381}

\title{
Fast(er) Exact Decoding and Global Training for Transition-Based Dependency Parsing via a Minimal Feature Set
}

\author{
    Tianze Shi\\
    Cornell University\\
    {\tt\small tianze@cs.cornell.edu} \\
    \And
    Liang Huang\\
    Oregon State University\\
    {\tt\small liang.huang.sh@gmail.com} \\
    \And
    Lillian Lee\\
    Cornell University\\
    {\tt\small llee@cs.cornell.edu} \\
}

\date{}

\def\basiceval#1{\the\numexpr#1\relax}

\newcommand{\specialcell}[2][c]{
    \begin{tabular}[#1]{@{}c@{}}#2\end{tabular}}

\newtoggle{comment}
\toggletrue{comment}

\mathchardef\mhyphen="2D %

\newcommand{\rootsym}{\ensuremath{\texttt{ROOT}}}
\newcommand{\shift}{\ensuremath{\mathsf{sh}}}

\newcommand{\rightarc}{\ensuremath{\mathsf{ra}}}
\newcommand{\reduce}{\ensuremath{\mathsf{re}}}

\newcommand{\reftab}[1]{Table~\ref{#1}}
\newcommand{\reffig}[1]{Figure~\ref{#1}}
\newcommand{\refsec}[1]{\S\ref{#1}}

\newcommand{\bivec}[1]
{\ensuremath{\stackrel{\raisebox{-.05cm}{\text{\resizebox{0.27cm}{0.08cm}{$\rightarrow\hspace{-0.01cm}\leftarrow$}}}}
                      {\vphantom{t}\smash#1}
             }}
\newcommand{\svec}[1]{\ensuremath{\bivec{s}_{#1}}}
\newcommand{\bvec}[1]{\ensuremath{\bivec{b}_{#1}}}
\newcommand{\wvec}[1]{\ensuremath{\bivec{w}_{#1}}}

\newcommand{\larc}{\ensuremath{^\curvearrowleft}}
\newcommand{\rarc}{\ensuremath{^\curvearrowright}}

\DeclareDocumentCommand{\consechalftriangles}{O{1.0} O{0.5} m m}{
    \begin{scope}[scale=0.9,thick]
        \draw[anchor=mid] (0, 0) node[above] {\ensuremath{#3}} -- ++({#2}, -0.5) -- ++(-{#2}, 0) -- cycle;
        \draw[anchor=mid] ({#1}, 0) node[above] {\ensuremath{#4}} -- ({#2}, -0.5) -- ({#1}, -0.5) -- cycle;
    \end{scope}
}

\DeclareDocumentCommand{\basicconsechalftriangles}{O{0.5} O{0.25} m m}{
    \begin{scope}[scale=0.9,thick]
        \draw[anchor=mid] (0, 0) node[above] {\ensuremath{#3}} -- ++({#2}, -0.25) -- ++(-{#2}, 0) -- cycle;
        \draw[anchor=mid] ({#1}, 0) node[above] {~~~\ensuremath{#4}} -- ({#2}, -0.25) -- ({#1}, -0.25) -- cycle;
    \end{scope}
}

\DeclareDocumentCommand{\trapezoid}{O{2.0} O{1.0} O{0.5} m m}{
    \begin{scope}[scale=0.9,thick]
        \draw[anchor=mid] (0, 0) node[above] {\ensuremath{#4}} -- (0, -{#2}) -- ({#1}, -{#2}) -- ({#1}, -{#3}) node [above] {\ensuremath{#5}} -- cycle;
    \end{scope}
}

\DeclareDocumentCommand{\righttriangle}{O{0.5} O{0.5} m m}{
    \begin{scope}[scale=0.9,thick]
        \draw[anchor=mid] (0, 0) node[above] {\ensuremath{#3}} -- (0, -{#2}) -- ({#1}, -{#2}) node [above right] {\ensuremath{#4}} -- cycle;
    \end{scope}
}

\DeclareDocumentCommand{\lefttriangle}{O{0.5} O{0.5} m m}{
    \begin{scope}[scale=0.9,thick]
        \draw[anchor=mid] (0, -{#2}) node[above left] {\ensuremath{#3}} -- ({#1}, -{#2}) -- ({#1}, 0) node [above] {\ensuremath{#4}} -- cycle;
    \end{scope}
}

\DeclareDocumentCommand{\standardtriangle}{O{0.5} O{0.25} O{0.25} m m m}{
    \begin{scope}[scale=0.9,thick]
        \draw[anchor=mid] (0, 0) node[left] {\ensuremath{#4}} -- ({#2}, {#3}) node [above] {\ensuremath{#5}} -- ({#1}, 0) node [right] {\ensuremath{#6}} -- cycle;
    \end{scope}
}

\newcommand{\lreduce}{\ensuremath{\mathsf{re_{\small \curvearrowleft}}}\xspace}
\newcommand{\rreduce}{\ensuremath{\mathsf{re_{\small \curvearrowright}}}\xspace}

\newtheorem*{theorem*}{Theorem}
\newtheorem*{lemma*}{Lemma}
\newtheorem{lemma}{Lemma}

\newenvironment{sproof}{%
 \proof}{\endproof}

 \newcommand{\terminal}{\tau}

\newcommand{\cost}{\ensuremath{\mathit{cost}}\xspace}
\newcommand{\head}{\ensuremath{\mathit{head}}\xspace}

\begin{document}

\maketitle

\begin{abstract}
We
first
present a minimal feature set for transition-based dependency parsing,
continuing
a recent trend started by \namecite{kiperwasser2016lstm} and \namecite{cross2016minimal}
of using
bi-directional LSTM features.
We plug our minimal feature set into the dynamic-programming framework of
\citet*{huang2010dp} and \citet{kuhlmann2011dp}
to produce the first
implementation of
worst-case
$O(n^3)$
exact decoders
for
arc-hybrid and arc-eager transition systems.
With
our minimal features, we also present $O(n^3)$ global training methods.
Finally, using ensembles
including our new parsers, we achieve the best
unlabeled attachment score
reported (to our knowledge) on the Chinese Treebank and
the ``second-best-in-class'' result on the English Penn Treebank.
\end{abstract}

\section{Introduction}
\label{sec:intro}

It used to be the case that the most accurate dependency parsers made global decisions and employed exact decoding.
But transition-based dependency parsers (TBDPs) have recently
achieved state-of-the-art performance,
despite
the fact
that for efficiency reasons, they are usually trained to make local,
rather than global,
decisions and the decoding process is done approximately,
rather than exactly
 \cite{weiss2015structured,dyer2015lstm,andor2016global}.
The key efficiency issue for decoding is as follows.
In order to make
accurate
(local) attachment decisions,
historically,
TBDPs have
required a large set of features
in order to access rich information about particular {\em positions} in the stack and buffer of the current parser
configuration.
But consulting many positions
means that although polynomial-time exact-decoding algorithms do exist, having been introduced by \citet{huang2010dp} and \citet{kuhlmann2011dp}, unfortunately, they are
prohibitively costly in practice, since the number of positions considered can factor into the exponent of the running
time.
For instance,
\citeauthor{huang2010dp}
employ a fairly reduced set of nine positions, but the worst-case running time for
the exact-decoding version of their algorithm is
$O(n^6)$ (originally reported as $O(n^7)$) for a length-$n$
sentence.
As an extreme case, \namecite{dyer2015lstm}
use an LSTM to summarize {\em arbitrary} information on the stack, which completely rules out dynamic programming.
Recently, \newcite{kiperwasser2016lstm} and \newcite{cross2016minimal} applied
bi-directional long short-term memory networks \cite[bi-LSTMs]{Graves} %
to derive feature representations for parsing,
because these networks capture wide-window contextual information well.
Collectively, these two sets of authors demonstrated that with bi-LSTMs,  {\em four} positional features suffice for
the arc-hybrid parsing system (K\&G), and {\em three} suffice for arc-standard (C\&H).\footnote{
We
note that K\&G were not focused on
minimizing positions,
although they explicitly noted the %
implications of doing so: ``While not
explored in this work, [fewer positions] results in very compact state signatures, [which is] very appealing for use in
transition-based parsers that employ dynamic-programming search'' (pg. 319).
C\&H also noted in their follow-up \cite{cross2016oracle} the possibility of future work using dynamic programming
thanks to simple features.}

Inspired by their work,
we arrive at a {\em minimal} feature set
for
arc-hybrid and arc-eager:
it contains {\em only
two} positional bi-LSTM vectors,
suffers almost no loss in performance in comparison to larger sets, and out-performs a single position.
(Details regarding the situation with arc-standard can be found in \refsec{sec:minimal}.)

Our minimal feature set
plugs into \citeauthor{huang2010dp}'s and \citeauthor{kuhlmann2011dp}'s dynamic programming framework
to produce the first
implementation of
$O(n^3)$
{\em exact} decoders for  arc-hybrid and arc-eager parsers.
We also enable  and implement
$O(n^3)$ {\em global} training methods.
Empirically, \emph{ensembles}
containing
our minimal-feature, globally-trained and exactly-decoded models
produce the best unlabeled
attachment score (UAS) reported (to our knowledge) on the Chinese Treebank
and
the ``second-best-in-class'' result on the English Penn Treebank.\footnote{
Our ideas were subsequently adapted to the {\em labeled} setting by
\citet*{shi+wu+chen+cheng:2017} in their submission to the CoNLL 2017 shared task on Universal Dependencies parsing. Their
team achieved the second-highest {\em labeled} attachment score in general and had the top average performance on the surprise languages.}

Additionally, we provide a slight update to the theoretical connections previously
drawn by \citet*{gomez2008deductive,gomez2011journal} between TBDPs and the {\em graph-based} dependency parsing
algorithms of \citet{eisner96} and \citet{eisner99},
including results regarding the arc-eager parsing system.

\section{A Minimal Feature Set}
\label{sec:minimal}

TBDPs incrementally process
a sentence
by making transitions through search states representing parser configurations.
Three of the main transition systems in use today (formal introduction in \refsec{sec:transition})
all maintain the following two data structures in their configurations:
(1) a stack of partially parsed subtrees  and (2) a buffer
(mostly)
of
unprocessed sentence tokens.
To featurize
 configurations
for
use in
a scoring function,
it is common to have features
that extract information about the first several elements
on the stack and the buffer, such as their word forms and part-of-speech (POS) tags.
We refer to these as {\em positional features},
as each feature relates to
a particular position
in the stack or buffer.
Typically, millions of
sparse
indicator features
(often developed via manual engineering)
are used.
In contrast, \citet{chenmanning2014} introduce
a feature set consisting of
\emph{dense}
word-, POS-, and dependency-label embeddings.  While dense, these features are for the same 18 positions that have been
typically used in prior work.
Recently,
\citet{kiperwasser2016lstm} and \citet{cross2016minimal}
adopt
bi-directional LSTMs, which have nice expressiveness and context-sensitivity properties, to
reduce
the number of positions considered down to four and three, for different transition systems, respectively.
This naturally begs the question, what is the lower limit on
the number of positional features necessary for a parser
to perform well?
\citet{kiperwasser2016lstm} reason that for the
{arc-hybrid} system,  the first and second
items on the stack and the first buffer
item --- denoted by $s_0$, $s_1$, and $b_0$, respectively --- are required; they additionally include
the third stack item, $s_2$,
because it may not be adjacent to the others in the original sentence.
For arc-standard, \citet{cross2016minimal} argue for the necessity of $s_0$, $s_1$, and $b_0$.

We address the lower-limit question empirically, and find that,
surprisingly,
{\em two positions
suffice}
for the greedy arc-eager and arc-hybrid parsers.
We also provide empirical support for \citeauthor{cross2016minimal}'s argument for the necessity of three features for arc-standard.
 In the
rest of
this
section,  we explain our
 experiments, run only on an {English development} set,  that support this conclusion;
the
results are depicted in Table \ref{tbl:minimal}.
We later explore the {\em implementation implications} in
\S \ref{sec:combined-notation+dp}-\ref{sec:decode-train}  and then  {\em test-set} parsing-accuracy
in \refsec{sec:exp}.

\bigskip

\label{sec:min-details}

\begin{table}[]
\centering
\vspace{6pt}
\small
\addtolength{\tabcolsep}{-.08cm} %
\begin{tabular}{@{\hspace{0.1em}}r@{\hspace{0.5em}}rcc@{\hspace{0.35em}}}
\toprule
        Features & Arc-standard & Arc-hybrid & Arc-eager \\
\midrule
    {$\{\svec{2}, \svec{1}, \svec{0}, \bvec{0}\}$}
            & $93.95_{\pm 0.12}$ & $94.08_{\pm 0.13}$ & $93.92_{\pm 0.04}$ \\
    $\{\svec{1},\svec{0},\bvec{0}\}$
            & $94.13_{\pm 0.06}$ & $94.08_{\pm 0.05}$ & $93.91_{\pm 0.07}$ \\ \cline{2-2}
    $\{\svec{0},\bvec{0}\}$
            & \multicolumn{1}{r|}{$54.47_{\pm 0.36}$} & $94.03_{\pm 0.12}$ & $93.92_{\pm 0.07}$ \\ \cline{3-4}
    $\{\bvec{0}\}$
            & $47.11_{\pm 0.44}$ & $52.39_{\pm 0.23}$ & $79.15_{\pm 0.06}$ \\
\bottomrule
\end{tabular}\\[0.2cm]
\begin{tabular}{@{\hspace{2.3em}}r@{\hspace{0.5em}}c@{\hspace{0.8em}}cc@{\hspace{0.75em}}}
\toprule
Min positions  & Arc-standard & Arc-hybrid & Arc-eager \\
\midrule
 K\&G 2016a    &  -           & 4          & - \\
 C\&H 2016a    & 3            & -          & -\\
 our work      & 3            & {\bf 2}          & {\bf 2}\\
\bottomrule
\end{tabular}
\caption{
{\em Top:} English PTB
dev-set %
UAS\%
for
progressively smaller sets of positional features,
for greedy parsers %
with different transition systems.
The ``double-arrow'' notation indicates vectors produced by a {\em bi}-directional LSTM.
Internal
lines highlight
large performance drop-offs when a feature is deleted.
{\em Bottom:} sizes of the minimal feature sets
in \namecite{kiperwasser2016lstm}, \namecite{cross2016minimal},
and our work.
}
\label{tbl:minimal}
\end{table}
\addtolength{\tabcolsep}{.05cm}

We
employ the same model architecture as \citet{kiperwasser2016lstm}.
Specifically,
we first use a bi-LSTM to encode
an $n$-token sentence, treated as a sequence of
per-token concatenations of
word- and POS-tag embeddings,
into a sequence of vectors
$[\wvec{1}, \ldots,\wvec{n}]$,
where each $\wvec{i}$ is the output of the bi-LSTM at time step $i$.
(The double-arrow
notation for these vectors emphasizes the
{bi}-directionality of their origin).
Then, for a given parser configuration, stack positions are represented by $\svec{j}$, defined as  $\wvec{i(s_j)}$
where $i(s_j)$ gives the position in the sentence of the token that is
the head
of the
tree in $s_j$.
Similarly, buffer positions  are represented by \bvec{j}, defined as
$\wvec{i(b_j)}$ for the token at buffer position $j$.
Finally, as in \newcite{chenmanning2014}, we use a multi-layer perceptron to score possible transitions from the given
configuration, where the input
is the concatenation of some selection of the \svec{j}s  and \bvec{k}s.
We use greedy decoders,  and train the models with dynamic oracles \cite{goldberg2013}.

\reftab{tbl:minimal} reports the parsing accuracy that results for feature sets of size four, three, two, and one for
three commonly-used transition systems.  The data is the development section of the English Penn Treebank (PTB), and
experimental settings are as described in our other experimental section, \refsec{sec:exp}.
We see that we can go down to three or, in the {arc-hybrid} and {arc-eager}  transition systems, even two
positions
with very
little loss in performance, but not further.
We
therefore
call $\{\svec{0},\bvec{0}\}$ our
{\em minimal} feature set
with respect to arc-hybrid and arc-eager, and empirically confirm that \citeauthor{cross2016minimal}'s $\{\svec{0}, \svec{1}, \bvec{0}\}$ is minimal for arc-standard; see Table~\ref{tbl:minimal} for a summary.\footnote{
We tentatively conjecture that the following might explain the observed phenomena, but stress that we don't currently see
a concrete way to test the following hypothesis.
With $\{\svec{0}, \bvec{0}\}$,
in the arc-standard case, situations can arise where there are
multiple possible transitions with missing information.  In contrast, in the arc-hybrid case, there is only one possible
transition with missing information (namely, \rreduce, introduced in \S \ref{sec:transition});
perhaps
$\svec{1}$
is therefore
not so crucial for arc-hybrid
in practice?
}

\section{Dynamic Programming for TBDPs
}
\label{sec:combined-notation+dp}
As stated in the introduction, our minimal feature set from  \refsec{sec:minimal} plugs into \citeauthor{huang2010dp}
and \citeauthor{kuhlmann2011dp}'s dynamic programming (DP) framework.  To help explain the connection, this section  provides
an overview of the DP framework.  We draw heavily from the presentation of \citet{kuhlmann2011dp}.

\subsection{Three Transition Systems}
\label{sec:transition}

Transition-based parsing \cite{nivre2008,kubler2009dependency}
is an incremental parsing framework
based on transitions between parser configurations.
For a sentence to be parsed, the system starts from a corresponding initial configuration,
and
attempts to sequentially apply transitions until
a configuration corresponding to
a full parse is produced.
Formally, a transition system is defined as
$\mathcal{S}=(C, T, c^s, C_\terminal)$,
where $C$ is a
nonempty
set of configurations,
each $t\in T : C\rightharpoonup C$ is a transition
function between configurations,
$c^s$ is an initialization function that maps an input sentence to an initial configuration,
and $C_\terminal\subseteq C$ is a set of terminal configurations.

All systems we consider share a common tripartite representation for configurations: when we write
$c=(\sigma, \beta, A)$ for some $c\in C$, we are referring to
a stack $\sigma$ of partially parsed subtrees; a buffer $\beta$ of
unprocessed
tokens and, optionally, at its
beginning, a subtree with only left descendants;
and a set $A$ of elements $(h, m)$, each of which is
an attachment (dependency arc) with head $h$ and modifier $m$.\footnote{
For simplicity, we only present unlabeled parsing here. See \citet{shi+wu+chen+cheng:2017} for labeled-parsing results.}
We write $m\larc h$ to indicate that $m$ left-modifies $h$, and $h\rarc m$
to indicate that $m$ right-modifies $h$.
For a sentence $w=w_1,...,w_n$,
the initial configuration is $(\sigma_0, \beta_0, A_0)$, where
$\sigma_0$ and  $A_0$ are empty and $\beta_0=[\rootsym|w_1,...,w_n]$;
$\rootsym$ is a special node denoting the root of the parse tree\footnote{
Other presentations place $\rootsym$ at the end of the buffer or omit it entirely \cite{ballesteros2013root}.}
(vertical bars are a notational convenience for indicating different parts of the buffer or stack; our convention is to depict the buffer first element leftmost, and to depict the stack first element rightmost).
All terminal configurations
have
an empty buffer and a stack containing only
$\rootsym$.

\paragraph{Arc-Standard}
The arc-standard system \cite{nivre2004standard} is motivated by bottom-up parsing:
each dependent has to be complete before being attached.
The three transitions, shift (\shift,
move a token from the buffer to the stack),
right-reduce (\rreduce,
reduce and attach a right modifier), and
left-reduce (\lreduce,
reduce and attach a left modifier),
are defined as:

\vspace{-16pt}
{\setlength{\mathindent}{0cm}
\begin{align*}
    &\shift\lbrack(\sigma, b_0|\beta, A)\rbrack=(\sigma|b_0, \beta, A)\\ %
    &\rreduce\lbrack(\sigma|s_1|s_0, \beta, A)\rbrack=(\sigma|s_1, \beta, A\cup\{(s_1,s_0)\})\\ %
    &\lreduce\lbrack(\sigma|s_1|s_0, \beta, A)\rbrack=(\sigma|s_0, \beta, A\cup\{(s_0,s_1)\}) %
\end{align*}}
\vspace{-16pt}

\noindent

\paragraph{Arc-Hybrid}
The arc-hybrid system \cite{ym03,gomez2008deductive,kuhlmann2011dp}
has the same definitions of $\shift$ and $\rreduce$ as arc-standard,
but forces the collection of left modifiers before right modifiers via its $b_0$-modifier $\lreduce$ transition.
This contrasts with arc-standard, where the attachment of left and right modifiers can be interleaved on the stack.

\vspace{-16pt}
{\setlength{\mathindent}{0cm}
\begin{align*}
    &\shift\lbrack(\sigma, b_0|\beta, A)\rbrack=(\sigma|b_0, \beta, A)\\ %
    &\rreduce\lbrack(\sigma|s_1|s_0, \beta, A)\rbrack=(\sigma|s_1, \beta, A\cup\{(s_1,s_0)\})\\ %
    &\lreduce\lbrack(\sigma|s_0, b_0|\beta, A)\rbrack=(\sigma, b_0|\beta, A\cup\{(b_0,s_0)\}) %
\end{align*}}
\vspace{-16pt}

\noindent

\paragraph{Arc-Eager}
In contrast to the former two systems, the arc-eager system \cite{nivre2003eager} makes attachments as early as possible --- even if
a modifier has not yet received all of its own modifiers.
This behavior is accomplished by decomposing the right-reduce transition into two independent transitions,
one making the attachment ($\rightarc$) and one reducing the right-attached child ($\reduce$).

\vspace{-16pt}
{\setlength{\mathindent}{0cm}
\begin{align*}
    &\shift\lbrack(\sigma, b_0|\beta, A)\rbrack=(\sigma|b_0, \beta, A)\\ %
    &\lreduce\lbrack(\sigma|s_0, b_0|\beta, A)\rbrack=(\sigma, b_0|\beta, A\cup\{(b_0,s_0)\})\\ %
            & \hspace{0.5cm} \mbox{(precondition: } s_0 \mbox{ not attached to any word)} \\ %
    &\rightarc\lbrack(\sigma|s_0, b_0|\beta, A)\rbrack=(\sigma|s_0|b_0, \beta, A\cup\{(s_0,b_0)\})\\ %
    &\reduce\lbrack(\sigma|s_0, \beta, A)\rbrack=(\sigma, \beta, A) \\%
              &  \hspace{0.5cm} \mbox{(precondition: } s_0 \mbox{ has been attached to its head)} %
\end{align*}}
\vspace{-16pt}

\subsection{Deduction and Dynamic Programming}
\label{sec:dp}

\begin{figure*}
    \small
    \begin{subfigure}[b]{0.5\textwidth}
        \centering
        \addtolength{\tabcolsep}{-.15cm}
        \begin{tabular}{llll}
            {\bf Axiom} & $[0, 0, 1]$ & $\tikz[baseline=0pt]{\standardtriangle[0.25][0][0.25]{0}{0}{1}}$\\
            \\
            \multicolumn{2}{l}{{\bf Inference Rules}} \\

            \shift & $\inferrule{[i,h,j]}{[j,j,j+1]}$ &
            $\inferrule{\tikz[baseline=0pt]{\standardtriangle[0.8][0.4][0.3]{i}{h}{j}}\hspace{23pt}}
            {\hspace{20pt}\tikz[baseline=-12pt]{\standardtriangle[0.25][0][0.25]{j}{j}{j+1}}}$ & $j \leq n$  \\

            \rreduce & $\inferrule{[i,h_1,k]\quad[k,h_2,j]}{[i,h_1,j]}$ &
            $\inferrule{\tikz[baseline=0pt]{\standardtriangle{i}{h_1}{k}}\hspace{-6pt}\tikz[baseline=0pt]{\standardtriangle{}{h_2}{j}}}
            {\tikz[baseline=-12pt]{\standardtriangle[1.5][0.25][0.35]{i}{h_1}{j}}}$ & $h_1 \rarc h_2$  \\

            \lreduce & $\inferrule{[i,h_1,k]\quad[k,h_2,j]}{[i,h_2,j]}$ &
            $\inferrule{\tikz[baseline=0pt]{\standardtriangle{i}{h_1}{k}}\hspace{-6pt}\tikz[baseline=0pt]{\standardtriangle{}{h_2}{j}}}
            {\tikz[baseline=-12pt]{\standardtriangle[1.5][1.25][0.35]{i}{h_2}{j}}}$ & $h_1 \larc h_2$  \\
            {\bf Goal} & $[0, 0, n+1]$ &
            \multicolumn{2}{l}{$\tikz[baseline=0pt]{\standardtriangle[1.8][0.0][0.5]{0}{0}{n+1}}$}

        \end{tabular}
        \caption{Arc-standard}
        \label{fig:arc-standard}
    \end{subfigure}
    \newcommand{\rulesep}{\unskip\ \vrule\ }
    \rulesep
    \begin{subfigure}[b]{0.5\textwidth}
        \centering
        \addtolength{\tabcolsep}{-.15cm}
        \begin{tabular}{llll}
            {\bf Axiom} & $[0, 1]$ & $\tikz[baseline=-5pt]{\basicconsechalftriangles{0}{1}}$\\
            \\
            \multicolumn{2}{l}{{\bf Inference Rules}} \\

            $\shift$ & $\inferrule{[i,j]}{[j,j+1]}$ &
            $\inferrule{\tikz[baseline=-12pt]{\consechalftriangles{i}{j}}\hspace{38pt}}
            {\hspace{38pt}\tikz[baseline=-12pt]{\basicconsechalftriangles{j}{j\!+\!1}}}$ & $j \leq n$  \\

            $\rreduce$ & $\inferrule{[k,i]\quad[i,j]}{[k,j]}$ &
            $\inferrule{\tikz[baseline=-12pt]{\consechalftriangles{k}{i}}\tikz[baseline=-12pt]{\consechalftriangles{i}{j}}}
            {\tikz[baseline=-12pt]{\consechalftriangles[2.4][1.9]{k}{j}}}$ & $k \rarc i$  \\

            $\lreduce$ & $\inferrule{[k,i]\quad[i,j]}{[k,j]}$ &
            $\inferrule{\tikz[baseline=-12pt]{\consechalftriangles{k}{i}}\tikz[baseline=-12pt]{\consechalftriangles{i}{j}}}
            {\tikz[baseline=-12pt]{\consechalftriangles[2.4][0.5]{k}{j}}}$ & $i \larc j$
            \vspace{14pt}
            \\

            {\bf Goal} & $[0, n+1]$ &
            \multicolumn{2}{l}{\hspace{0.2pt} $\tikz[baseline=-12pt]{\consechalftriangles[2.4][2.3]{0}{n+1}}$}
        \end{tabular}
        \vspace{10pt}
        \caption{Arc-hybrid}
        \label{fig:arc-hybrid}
    \end{subfigure}
    \hrule %
    \begin{subfigure}[b]{0.5\textwidth}
        \centering
        \vspace{5pt}
        \addtolength{\tabcolsep}{-.1cm}
        \begin{tabular}{llll}
            {\bf Axiom} & $[0^0, 1]$ & $\tikz[baseline=-5pt]{\basicconsechalftriangles{0^0}{1}}$\\
            \\
            \multicolumn{2}{l}{{\bf Inference Rules}} \\

            $\shift$ & $\inferrule{[i^b,j]}{[j^0,j+1]}$ &
            $\inferrule{\tikz[baseline=-12pt]{\consechalftriangles{i^b}{j}}\hspace{38pt}}
            {\hspace{38pt}\tikz[baseline=-12pt]{\basicconsechalftriangles{j^0}{j\!+\!1}}}$ & $j \leq n$  \\

            $\rightarc$ & $\inferrule{[i^b,j]}{[j^1,j+1]}$ &
            $\inferrule{\tikz[baseline=-12pt]{\consechalftriangles{i^b}{j}}\hspace{38pt}}
            {\hspace{38pt}\tikz[baseline=-12pt]{\basicconsechalftriangles{j^1}{j\!+\!1}}}$ & \hspace{-5pt}\specialcell{$i\rarc j$ \\ $j \leq n$}  \\

            $\lreduce$ & $\inferrule{[k^b,i]\quad[i^0,j]}{[k^b,j]}$ &
            $\inferrule{\tikz[baseline=-12pt]{\consechalftriangles{k^b}{i}}\tikz[baseline=-12pt]{\consechalftriangles{i^0}{j}}}
            {\tikz[baseline=-12pt]{\consechalftriangles[2.4][0.5]{k^b}{j}}}$ & $i \larc j$ \\

            $\reduce$ & $\inferrule{[k^b,i]\quad[i^1,j]}{[k^b,j]}$ &
            $\inferrule{\tikz[baseline=-12pt]{\consechalftriangles{k^b}{i}}\tikz[baseline=-12pt]{\consechalftriangles{i^1}{j}}}
            {\tikz[baseline=-12pt]{\consechalftriangles[2.4][1.9]{k^b}{j}}}$ & $ $ \\

            {\bf Goal} & $[0^0, n+1]$ &
            \multicolumn{2}{l}{\hspace{0pt} $\tikz[baseline=-12pt]{\consechalftriangles[2.4][2.3]{0^0}{n+1}}$}
        \end{tabular}
        \caption{Arc-eager}
        \label{fig:arc-eager}
    \end{subfigure}
    \hspace*{.005cm}\rulesep
    \begin{subfigure}[b]{0.5\textwidth}
        \centering
        \vspace{5pt}
        \begin{tabular}{lll}
            {\bf Axioms} &
            \multicolumn{2}{l}{
            $\tikz[baseline=-7pt]{\righttriangle[0.2][0.3]{i}{i+1}}$ \quad
            $\tikz[baseline=-7pt]{\lefttriangle[0.2][0.3]{j}{j}}$ \quad $0\leq i, j\leq n$
            }\\
            \\
            \multicolumn{2}{l}{{\bf Inference Rules}} \\

            $\mathsf{right\mhyphen attach}$ &
            $\inferrule{\tikz[baseline=-15pt]{\trapezoid[0.9][0.6][0.2]{k}{i}}\tikz[baseline=-10pt]{\righttriangle[0.5][0.4]{i}{j}}}
            {\tikz[baseline=-12pt]{\righttriangle[1.75][0.6]{k}{j}}}$ \\
            $\mathsf{right\mhyphen reduce}$ &
            $\inferrule{\tikz[baseline=-9pt]{\righttriangle[0.5][0.4]{i}{k}}\tikz[baseline=-9pt]{\lefttriangle[0.5][0.4]{k}{j}}}
            {\tikz[baseline=-12pt]{\trapezoid[1.85][0.6][0.3]{i}{j}}}$ & $i\rarc j$ \\

            $\mathsf{left\mhyphen attach}$ &
            $\inferrule{\hspace{-4pt}\tikz[baseline=-10pt]{\lefttriangle[0.5][0.4]{k}{i}}\tikz[baseline=-10pt]{\trapezoid[0.9][0.4][-0.2]{i}{j}}}
            {\hspace{-4pt}\tikz[baseline=-12pt]{\lefttriangle[1.75][0.6]{k}{j}}}$ \\
            $\mathsf{left\mhyphen reduce}$ &
            $\inferrule{\tikz[baseline=-9pt]{\righttriangle[0.5][0.4]{i}{k}}\tikz[baseline=-9pt]{\lefttriangle[0.5][0.4]{k}{j}}}
            {\tikz[baseline=-12pt]{\trapezoid[1.85][0.3][-0.3]{i}{j}}}$ & $i\larc j$  \\

            {\bf Goal} &
            \multicolumn{2}{l}{$\tikz[baseline=0pt]{\standardtriangle[1.8][0.0][0.5]{}{0}{n+1}}$}

        \end{tabular}
        \caption{Edge-factored graph-based parsing.}
        \label{fig:edge-factored}
    \end{subfigure}
    \caption{
    \ref{fig:arc-standard}-\ref{fig:arc-eager}: \citeauthor{kuhlmann2011dp}'s inference rules for three transition systems, together
    with CKY-style visualizations of the local structures involved and, to their right, conditions for the rule to apply.  \ref{fig:edge-factored}: the edge-factored graph-based parsing algorithm \cite{eisner99} discussed in \refsec{sec:connections}.
    }
    \label{fig:deduction}
\end{figure*}

\newcite{kuhlmann2011dp} reformulate the three transition systems just discussed as deduction
systems \cite{pereira+warren:83a,shieber+schabes+pereira:95a}, wherein transitions serve as inference rules;
these are given as the lefthand sides of the first three subfigures in \reffig{fig:deduction}.
For a given $w=w_1,...,w_n$, assertions take the form $[i, j, k]$ (or, when applicable, a two-index shorthand to be discussed soon),
meaning that there exists a sequence of transitions that, starting from a configuration wherein
$\head(s_0)=w_i$, results in
an ending configuration wherein
$\head(s_0)=w_j$ and $\head(b_0)=w_k$.
If we define $w_0$ as $\rootsym$ and $w_{n+1}$ as an end-of-sentence marker, then the goal theorem can be stated as
$[0,0,n+1]$.

For arc-standard, we depict an assertion  $[i, h, k]$ as a subtree whose root (head) is the token at $h$.
Assertions of the form $[i, i, k]$ play an important role for
arc-hybrid and arc-eager, and we employ the special shorthand $[i,k]$ for them in \reffig{fig:deduction}.
In that figure, we also graphically depict such situations as two consecutive half-trees
with roots $w_i$ and $w_k$, where all tokens between $i$ and $k$ are already attached.
The superscript $b$ in an arc-eager assertion $[i^b,j]$
is
an indicator variable for whether $w_i$ has been attached to its head
($b=1$) or not ($b=0$)
after the transition sequence is applied.

\citet{kuhlmann2011dp} show that all three deduction systems can be directly ``tabularized'' and dynamic programming (DP) can
be applied, such that,
{\em ignoring for the moment the issue of incorporating complex features} (we return to this later), time and space needs are
low-order polynomial.
Specifically,
as the two-index shorthand $[i,j]$ suggests, arc-eager and arc-hybrid systems
can be implemented to take
$O(n^2)$ space and $O(n^3)$ time;
the arc-standard system requires $O(n^3)$ space and $O(n^4)$ time
(if one applies the so-called hook trick \cite{eisner99}).

Since an $O(n^4)$  running time is not sufficiently practical even in the simple-feature case, {\em in the remainder of this paper we consider only the arc-hybrid and arc-eager systems, not  arc-standard.}

\section{Practical Optimal Algorithms Enabled By Our Minimal Feature Set}
\label{sec:decode-train}

Until now, no one had suggested a set of positional features that was both information-rich
enough for accurate parsing {\em and} small enough to obtain the $O(n^3)$ running-time promised above.  Fortunately, our bi-LSTM-based $\{\svec{0},\bvec{0}\}$ feature set qualifies, and enables the fast optimal procedures described in this section.

\subsection{Exact Decoding}
\label{sec:decoding}
Given an input sentence, a TBDP must choose among a potentially exponential number of corresponding transition sequences.
We assume access to functions $f_t$ that score individual configurations, where these functions are indexed
by the transition functions $t \in T$.  For a fixed transition sequence $\mathbf{t}  = t_1, t_2, \ldots$, we
use $c_i$ to denote the configuration that results after applying $t_i$.

Typically, for efficiency reasons, greedy left-to-right decoding is employed: the
next transition $t_i^*$
out of $c_{i-1}$ is
$\arg\max_{t}f_t(c_{i-1})$, so that past and future decisions are not taken into account.
The score $F(\mathbf{t})$ for the
transition
 sequence is induced by summing the relevant $f_{t_i}(c_{i-1})$ values.

However, our use of minimal feature sets enables direct computation of an argmax over the entire space of transition sequences,
$\arg\max_{\mathbf{t}}F(\mathbf{t})$,
via dynamic programming, because
our positions don't rely on
any information ``outside'' the deduction rule indices,
thus eliminating the need for additional state-keeping.

We show how to integrate the scoring functions
for
the arc-eager system;
the arc-hybrid system is handled similarly.
The score-annotated rules are as follows:
\vspace{-5pt}
\begin{equation*}
    \inferrule{[i^b,j]:v}{[j^0,j+1]:0}(\shift)~~
    \inferrule{[k^b,i]:v_1\quad[i^0,j]:v_2}{[k^b,j]:v_1+v_2+\Delta}(\lreduce)
\end{equation*}
\vspace{-5pt}

\noindent where $\Delta=f_{\shift}(\wvec{k},\wvec{i})+f_{\lreduce}(\wvec{i},\wvec{j})$
--- abusing
notation by
referring
to configurations by their features.
The left-reduce rule says that we
can
first take the sequence of transitions
asserted by $[k^b,i]$,
which has a score of $v_1$,
and
then a shift transition moving $w_i$ from $b_0$ to $s_0$.
This means that
the initial condition for $[i^0,j]$ is met, so
we can take the sequence of transitions asserted by $[i^0,j]$ --- say it has score $v_2$ ---
and finally a left-reduce transition to finish composing the larger transition sequence.
Notice that the scores for \shift\ and \rightarc\ are $0$,
as the scoring of
these transitions is
accounted for by reduce rules elsewhere in the sequence.

\subsection{Global Training}
\label{sec:training}
We employ large-margin training
that considers each transition sequence globally.
Formally, for a training sentence
$w=w_1, \ldots, w_n$ with gold transition sequence
\newcommand{\gold}{\mathbf{t^{{\rm gold}}}}
$\gold$,
our loss function is

\vspace{-12pt}
\hspace{-20pt}
    $$\max_{
    {\mathbf t}}
    \left(
    {F(\mathbf{t})+\cost(\mathbf{t^{{\rm gold}}}, \mathbf{t})}    -F(\gold)
    \right)
    $$
\noindent where $\cost(\mathbf{t^{{\rm gold}}}, \mathbf{t})$
is a custom margin for
taking $\mathbf{t}$ instead of $\mathbf{t^{{\rm gold}}}$ ---
specifically, the number of mis-attached nodes.
Computing this max
can again be done efficiently with a slight modification to the scoring of reduce transitions:

\vspace{-5pt}
\begin{equation*}
    \inferrule{[k^b,i]:v_1~~[i^0,j]:v_2}{[k^b,j]:v_1+v_2+\Delta'}(\lreduce)
\end{equation*}
where $\Delta'=\Delta + {\mathbf 1}\left({\head(w_i)\neq w_j}\right)$.
This
loss-augmented inference or cost-augmented decoding \cite{taskar2005structured,smith2011book} technique
has previously been applied to graph-based parsing by \citet{kiperwasser2016lstm}.

\paragraph{Efficiency Note}
The computation decomposes into two parts:
scoring all feature combinations, and using DP to find a proof for the goal theorem in the deduction system.
Time-complexity analysis
is usually given in terms of the latter,
but the
former might have a large constant factor,
such as $10^4$ or worse for neural-network-based scoring functions.
As a result, in practice,
with a small $n$,
scoring with the feature set $\{\svec{0},\bvec{0}\}$ ($O(n^2)$)
can be as time-consuming as the decoding steps ($O(n^3)$)
for the arc-hybrid and arc-eager systems.

\noindent

\begin{table*}[!ht]
\centering
\begin{tabular}{ccc|c@{\hspace{0.5em}}c|c@{\hspace{0.5em}}c}
    \toprule
    \multirow{2}{*}{Model} & \multirow{2}{*}{Training} & \multirow{2}{*}{Features} & \multicolumn{2}{c}{PTB} & \multicolumn{2}{c}{CTB} \\
                            &                          &                           & UAS (\%) & UEM (\%)     & UAS (\%) & UEM (\%)     \\
    \midrule
    {Arc-standard} & Local     & {\small $\{\svec{2}, \svec{1}, \svec{0}, \bvec{0}\}$} &
        $93.95_{\pm 0.12}$ &  $52.29_{\pm 0.66}$ &  $88.01_{\pm 0.26}$ &  $36.87_{\pm 0.53}$\\
    \midrule
    \multirow{3}{*}{Arc-hybrid}  & Local     & {\small $\{\svec{2}, \svec{1}, \svec{0}, \bvec{0}\}$} &
        $93.89_{\pm 0.10}$ &  $50.82_{\pm 0.75}$ &  $87.87_{\pm 0.17}$ &  $35.47_{\pm 0.48}$\\
                                & Local     & {\small $\{\svec{0}, \bvec{0}\}$} &
        $93.80_{\pm 0.12}$ &  $49.66_{\pm 0.43}$ &  $87.78_{\pm 0.09}$ &  $35.09_{\pm 0.40}$\\
                                & Global    & {\small $\{\svec{0}, \bvec{0}\}$} &
        $94.43_{\pm 0.08}$ &  $53.03_{\pm 0.71}$ &  $88.38_{\pm 0.11}$ &  $36.59_{\pm 0.27}$\\
    \midrule
    \multirow{3}{*}{Arc-eager}  & Local     & {\small $\{\svec{2}, \svec{1}, \svec{0}, \bvec{0}\}$} &
        $93.80_{\pm 0.12}$ &  $49.66_{\pm 0.43}$ &  $87.49_{\pm 0.20}$ &  $33.15_{\pm 0.72}$\\
                                & Local     & {\small $\{\svec{0}, \bvec{0}\}$} &
        $93.77_{\pm 0.08}$ &  $49.71_{\pm 0.24}$ &  $87.33_{\pm 0.11}$ &  $34.17_{\pm 0.41}$\\
                                & Global    & {\small $\{\svec{0}, \bvec{0}\}$} &
        ${\bf 94.53}_{\pm 0.05}$ &  $53.77_{\pm 0.46}$ &  ${\bf 88.62}_{\pm 0.09}$ &  ${\bf 37.75}_{\pm 0.87}$\\
    \midrule
        Edge-factored            & Global     & {\small $\{\bivec{h}, \bivec{m}\}$} &
        $94.50_{\pm 0.13}$ &  ${\bf 53.86}_{\pm 0.78}$ &  $88.25_{\pm 0.12}$ &  $36.42_{\pm 0.52}$\\
    \bottomrule
\end{tabular}
\caption{Test set performance
for different  training regimes and feature sets.
The models use the same decoders for testing and training.
For each setting,
the average and standard deviation across 5 runs with different random initializations are reported.
Boldface: best (averaged) result
per dataset/measure.
}
\label{tab:main-result}

\end{table*}

\section{Theoretical Connections}
\label{sec:connections}
Our minimal feature set brings implementation of practical optimal algorithms to TBDPs,
whereas previously only {\em graph-based} dependency parsers (GBDPs) ---  a radically different, non-incremental paradigm --- enjoyed the ability to deploy them.
Interestingly, for both the transition- and graph-based paradigms, the optimal algorithms build dependency trees bottom-up
from local structures.  It is thus natural to wonder if there are deeper, more formal connections between the two.

In previous work,
\citet{kuhlmann2011dp}
related the arc-standard system
to the classic CKY algorithm
\cite{CKY-C,CKY-K,CKY-Y}
in a manner clearly suggested by \reffig{fig:arc-standard};
CKY can be viewed as a very simple graph-based approach.
\citet{gomez2008deductive,gomez2011journal} formally prove that
sequences of steps in the {\em edge-factored} GBDP \citep{eisner96} can be used to emulate
any individual step in the arc-hybrid system \cite{ym03} and the \citet[\reffig{fig:edge-factored}]{eisner99} version.
However, they did not draw an explicitly direct connection between \citet{eisner99}
and TBDPs.

Here, we provide an update to these
previous findings, stated in terms of the expressiveness of
scoring functions, considered as
parameterization.

For
the edge-factored GBDP,
we write the score for an edge as $f_G(\bivec{h}, \bivec{m})$,
where $h$ is the head and $m$ the modifier.
A tree's score is the sum of its edge scores.
We say that a parameterized dependency parsing model A {\em contains} model B
if for every instance of parameterization in model B,
there exists an instance of model A
such that the two models assign the same score to every parse tree.
We claim:
\begin{lemma}
The arc-eager model presented in \refsec{sec:decoding}
contains
the edge-factored model.
\end{lemma}
\begin{sproof}
\vspace*{-.2cm}
Consider a given
edge-factored GBDP parameterized by $f_G$.
For any parse tree,
every edge $i\larc j$ involves two deduction rules,
and their contribution to the score of the final proof is
$f_{\shift}\mbox{(\wvec{k},\wvec{i})}
+f_{\lreduce}(\wvec{i},\wvec{j})$.
We set $f_{\shift}\mbox{(\wvec{k},\wvec{i})}=0$ and $f_{\lreduce}(\wvec{i},\wvec{j})=f_G(\wvec{j},\wvec{i})$.
Similarly, for edges  $k\rarc i$ in the other direction,
we set
$f_{\rightarc}\mbox{(\wvec{k},\wvec{i})}=f_G(\wvec{k},\wvec{i})$
and
$f_{\reduce}(\wvec{i},\wvec{j})=0$.
The parameterization we arrive at emulates exactly the scoring model of $f_G$.
\end{sproof}

We further claim that the arc-eager model is more expressive than
not only the edge-factored GBDP, but also the arc-hybrid model in our paper.

\begin{lemma}
The arc-eager model
contains
the arc-hybrid model.
\end{lemma}
\begin{sproof}
\vspace*{-.2cm}
We leverage the fact
that the arc-eager model divides the \shift\ transition in the arc-hybrid model
into two separate transitions, \shift\ and \rightarc.
When we constrain the parameters $f_{\shift} = f_{\rightarc}$ in the arc-eager model,
the model hypothesis space becomes exactly the same as arc-hybrid's.
\end{sproof}

The extra expressiveness of the arc-eager model comes from the scoring functions
$f_{\shift}$ and $f_{\reduce}$
that capture structural contexts other than head-modifier relations.
Unlike traditional higher-order graph-based parsing that directly models relations
such as siblinghood \cite{mcdonald2006sibling} or grandparenthood \cite{carreras2007second}, however,
the arguments in those two functions do not have any fixed type of structural interactions.

\section{Experiments}
\label{sec:exp}

\begin{figure*}[!ht]
\centering
\includegraphics[width=0.85\textwidth]{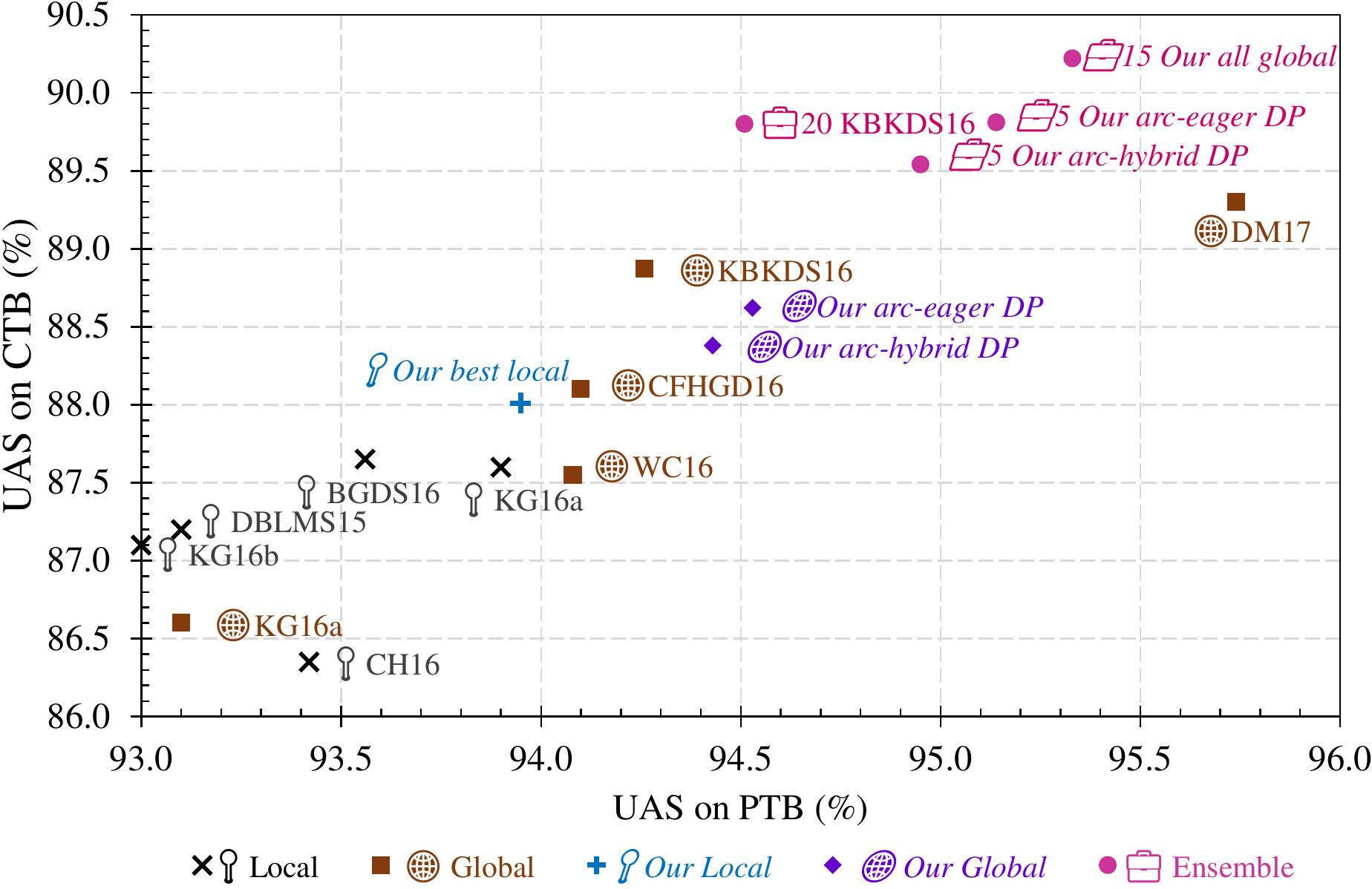}
\caption{
Comparing our UAS results with results from the literature. x-axis:  PTB;
 y-axis:  CTB.
Most datapoint labels give author initials and publication year; citations are in the bibliography.
Ensemble datapoints
are annotated with ensemble size.
\citet{weiss2015structured} and \citet{andor2016global} achieve
UAS of 94.26 and 94.61 on PTB with beam search, but did not report CTB results, and are
therefore omitted.}
        \nocite{kiperwasser2016lstm}
        \nocite{wang2016graph}
        \nocite{kuncoro2016ensemble}
        \nocite{cheng2016}
        \nocite{dozat2017biaffine}
        \nocite{kiperwasser2016lstm}
        \nocite{dyer2015lstm}
        \nocite{weiss2015structured}
        \nocite{andor2016global}
        \nocite{ballesteros2016exploration}
        \nocite{cross2016minimal}

\label{fig:sota-comparison}
\end{figure*}

\paragraph{Data and Evaluation}
We
experimented with English and Chinese.
For English, we used the Stanford Dependencies \cite{de2008stanford}
conversion
(via the Stanford parser 3.3.0) of the Penn Treebank \cite[PTB]{marcus1993ptb}.
As is standard,
we used \S 2-21 of the Wall Street Journal for training,
\S 22 for development, and \S 23 for testing; POS tags were predicted using 10-way jackknifing with the
Stanford max entropy tagger \cite{toutanova2003tagger}.
For Chinese, we used the Penn Chinese Treebank 5.1 \cite[CTB]{ctb},
with the same splits and head-finding rules for conversion to dependencies as \citet{zhang2008}.
We adopted the CTB's gold-standard tokenization and POS tags.
We report unlabeled attachment score (UAS) and sentence-level unlabeled exact match (UEM).
Following prior work, all punctuation is excluded from evaluation.
For each model, we initialized the network parameters with 5 different random seeds
and report performance average and standard deviation.

\paragraph{Implementation Details}
Our model structures
reproduce those of
\citet{kiperwasser2016lstm}.
We use 2-layer bi-directional LSTMs with 256 hidden cell units.
Inputs
are concatenations of 28-dimensional randomly-initialized part-of-speech embeddings
and 100-dimensional word vectors initialized from
GloVe
vectors \cite{glove} (English)
and  pre-trained
skip-gram-model vectors \cite{mikolov2013word2vec} (Chinese).
The concatenation of the bi-LSTM feature vectors is passed through a multi-layer perceptron (MLP) with 1 hidden layer
which has 256 hidden units and activation function tanh.
We set the dropout rate for the bi-LSTM \cite{gal2016dropout} and MLP \cite{dropout}
for each model
according to
development-set performance.%
\footnote{For bi-LSTM input and recurrent connections, we consider dropout rates
in $\{0., 0.2\}$,
and for MLP, $\{0., 0.4\}$.}
All parameters except the word embeddings are initialized uniformly \cite{glorot2010init}.
Approximately 1,000 tokens form a mini-batch for sub-gradient computation.
We train each model for 20 epochs and perform model selection based on development UAS.
The proposed structured loss function is optimized via Adam
\cite{kingma2015adam}.
The neural network computation is based on
the python interface to
DyNet \cite{dynet},
and the exact decoding algorithms are implemented in Cython.\footnote{
    See \scriptsize{\url{https://github.com/tzshi/dp-parser-emnlp17}} .
}

\paragraph{Main Results}

We implement exact decoders for the arc-hybrid and arc-eager systems,
and present the test performance of different model configurations in \reftab{tab:main-result},
comparing global models with local models.
All models use the same decoder for testing as during the training process.
Though no global decoder for the arc-standard system has been explored in this paper,
its local models are listed for comparison.
We also include an edge-factored graph-based model,
which is conventionally trained globally.
The edge-factored model scores bi-LSTM features for each head-modifier pair;
a maximum spanning tree algorithm is used to find the tree with the highest sum of edge scores.
For this model, we use \citeauthor{dozat2017biaffine}'s \shortcite{dozat2017biaffine} biaffine scoring model,
although in our case the model size is smaller.\footnote{The
same architecture and model size as other transition-based global models is used for fair comparison.}

Analogously to the dev-set results given in \refsec{sec:minimal}, on the test data,
the minimal feature sets perform as well as larger ones in locally-trained models.
And there exists a clear trend of global models outperforming local models
for the two different transition systems on both datasets.
This
illustrates the effectiveness of exact decoding and global training.
Of the three types of global models, the arc-eager arguably has the edge,
an empirical finding resonating with our theoretical comparison of their model expressiveness.

\paragraph{Comparison with State-of-the-Art Models}

\reffig{fig:sota-comparison} compares our algorithms' results with those of the state-of-the-art.\footnote{
We
exclude \citet{choe2016}, \citet{kuncoro2017rnng} and \citet{liu2017inorder}, which
convert constituent-based parses to dependency parses.
They produce higher PTB UAS,
but access more training information and do not
directly apply to datasets without constituency annotation.
}
Our models are competitive
and an ensemble of 15 globally-trained models (5 models each for arc-eager DP, arc-hybrid DP and edge-factored) achieves
95.33 and 90.22 on PTB and CTB, respectively, reaching the highest reported UAS on the CTB dataset,
and the second highest reported on the PTB dataset among dependency-based approaches.
\section{Related Work Not Yet Mentioned}
\label{sec:related}

\paragraph{Approximate Optimal Decoding/Training}
Besides dynamic programming
\cite{huang2010dp,kuhlmann2011dp},
various other approaches have been proposed for approaching global training and exact decoding.
Best-first and A* search \cite{klein2003best,sagae2006best,sagae2007best,zhao2013best,thang2015best,lee2016global}
give optimality certificates when %
solutions are found,
but have the same worst-case time complexity as the original search framework.
Other common approaches
to search a larger space at training or test time
include beam search \cite{zhang2011beam},
dynamic oracles \cite{goldberg2012oracle,goldberg2013,cross2016oracle} and error states \cite{vaswani2016error}.
Beam search records the
$k$ best-scoring transition prefixes to delay local hard decisions,
while the latter two leverage configurations deviating from the gold transition path
during training
to better simulate %
the test-time environment.

\paragraph{Neural Parsing}
Neural-network-based models are widely used in state-of-the-art dependency parsers
\cite{henderson2003nn,henderson2004nn,chenmanning2014,weiss2015structured,andor2016global,dozat2017biaffine}
because of %
their expressive representation power.
Recently,
\citet{stern2017minimal} have proposed minimal span-based features for constituency parsing.

Recurrent and recursive neural networks can be used to build representations that
encode complete configuration information or the entire parse tree \cite{le2014rnn,dyer2015lstm,kiperwasser2016rnn},
but these models cannot be readily combined with DP approaches, because
their state spaces cannot be merged into smaller sets and thus remain exponentially large.

\section{Concluding Remarks}

In this paper,
we have shown the following.
\begin{itemize}
\item The bi-LSTM-powered feature set $\{\svec{0},\bvec{0}\}$
is minimal yet highly effective %
for
arc-hybrid and arc-eager
transition-based parsing.
\item Since
DP %
algorithms for exact decoding \cite{huang2010dp,kuhlmann2011dp} have a run-time
dependence on the number of positional features, using our mere two effective positional features
results in a running time of $O(n^3)$, feasible for practice.
\item Combining exact decoding with global training --- which is also enabled by our minimal feature set ---
with an ensemble of parsers
 achieves
90.22 UAS on the Chinese Treebank and
95.33 UAS on the Penn Treebank: these are, to our knowledge, the best and second-best results to date on these data
sets among ``purely'' dependency-based approaches.
\end{itemize}

There are many directions for further exploration.
Two possibilities are to
create even better training methods, and to
find some way to extend our run-time improvements to
other transition systems.
It would also be interesting to further
investigate relationships between graph-based and dependency-based parsing.
In \refsec{sec:connections} we have mentioned important earlier work in this regard,
and provided an update to those formal findings.

In our work, we have brought exact decoding, which was formerly the province solely of
graph-based  parsing, to the transition-based paradigm. We hope that the future will bring more inspiration from an
integration of the two perspectives.

\paragraph*{Acknowledgments: an author-reviewer success story}
We sincerely thank all the reviewers for their extraordinarily careful and helpful comments.
Indeed, this paper originated as a short paper submission by TS\&LL to ACL 2017,
where an anonymous reviewer explained in
the review comments how, among other things, the DP run-time could be improved from $O(n^4)$ to $O(n^3)$.
In their author response, TS\&LL invited the reviewer to co-author, suggesting that they ask the conference organizers
to make the connection between anonymous reviewer and anonymous authors.  All three of us are truly grateful to PC co-chair
Regina Barzilay for implementing this idea, bringing us together!

We also thank Kai Sun for help with
Chinese word vectors,
and  Xilun Chen, Yao Cheng, Dezhong Deng, Juneki Hong,
Jon Kleinberg, Ryan McDonald, Ashudeep Singh, and Kai Zhao for discussions and suggestions.
TS and LL were supported in part by a
Google focused research grant to Cornell University.
LH was supported in part by
NSF IIS-1656051,
DARPA N66001-17-2-4030,
and a Google Faculty Research Award.

\bibliographystyle{emnlp_natbib}

\end{document}